\documentclass[journal]{IEEEtran}
\usepackage{blindtext}
\usepackage{times}
\usepackage{epsfig}
\usepackage{amssymb}
\usepackage{amsthm}
\usepackage{graphicx}
\usepackage{amsfonts}
\usepackage{amsmath}
\usepackage{array}
\usepackage{mdwmath}
\usepackage{mdwtab}
\usepackage{eqparbox}
\usepackage{fixltx2e}
\usepackage{stfloats}
\usepackage{url}
\usepackage{diagbox}
\usepackage{verbatim}
\usepackage{booktabs}
\usepackage{multirow}
\usepackage{bm}
\usepackage{mathtools}
\usepackage{breqn}
\usepackage{float}
\usepackage{tcolorbox}
\usepackage{todonotes}
\usepackage{pgfmath}
\usepackage{siunitx}
\usepackage{adjustbox}
\usepackage{threeparttable}
\usepackage{hyperref}
\usepackage{flushend}
\usepackage{soul}

\usepackage[font=footnotesize,labelfont=bf]{caption}
\usepackage{subcaption}

\definecolor{Gray}{gray}{0.85}
\definecolor{LightGray}{gray}{0.9}

\usepackage{algorithmic}
\usepackage[ruled, vlined, linesnumbered]{algorithm2e}

\usepackage{amsmath,amsfonts,bm}

\def\eqref#1{equation~\ref{#1}}

\def\1{\bm{1}}

\def\vu{{\bm{u}}}

\def\vy{{\bm{y}}}
\def\vz{{\bm{z}}}

\def\mI{{\bm{I}}}

\def\mU{{\bm{U}}}
\def\mV{{\bm{V}}}
\def\mW{{\bm{W}}}
\def\mX{{\bm{X}}}
\def\mY{{\bm{Y}}}

\DeclareMathAlphabet{\mathsfit}{\encodingdefault}{\sfdefault}{m}{sl}
\SetMathAlphabet{\mathsfit}{bold}{\encodingdefault}{\sfdefault}{bx}{n}

\ifCLASSINFOpdf
\else
\fi

\begin{document}
%
\title{Efficient Deep Spiking Multi-Layer Perceptrons with Multiplication-Free Inference}
%
%
%

\author{Boyan Li,
        Luziwei Leng,
        Shuaijie Shen,
        Kaixuan Zhang, 
        Jianguo Zhang,~\IEEEmembership{Senior Member,~IEEE}\\
       Jianxing Liao, 
       and Ran Cheng,~\IEEEmembership{Senior Member,~IEEE}
        \thanks{This work was supported in part by National Key Research and Development Program of China under Grant 2021YFF1200800 and National Natural Science Foundation of China under Grant 62276121. \emph{(Boyan Li and Luziwei Leng contributed equally to this work.)  (Corresponding author: Ran Cheng.) }
        }
        \thanks{Boyan Li, Shuaijie Shen, Kaixuan Zhang, Jianguo Zhang, and Ran Cheng are with the Department of Computer Science and Engineering, Southern University of Science and Technology, Shenzhen 518055, China
                (e-mail: liboyan2023@outlook.com; ranchengcn@gmail.com).
                }
        \thanks{Luziwei Leng and Jianxing Liao are with the Advanced Computing and Storage Lab, Huawei Technologies Co., Ltd., Shenzhen 518055, China (e-mail: \{lengluziwei; liaojianxing\}@huawei.com).}
}

\markboth{IEEE Transactions on Neural Networks and Learning Systems,~Vol.~, No.~, month~year}%
{Shell \MakeLowercase{\textit{et al.}}: Bare Demo of IEEEtran.cls for IEEE Journals}

\maketitle

\begin{abstract}
Advancements in adapting deep convolution architectures for Spiking Neural Networks (SNNs) have significantly enhanced image classification performance and reduced computational burdens. However, the inability of Multiplication-Free Inference (MFI) to align with attention and transformer mechanisms, which are critical to superior performance on high-resolution vision tasks, imposing limitations on these gains. To address this, our research explores a new pathway, drawing inspiration from the progress made in Multi-Layer Perceptrons (MLPs). We propose an innovative spiking MLP architecture that uses batch normalization to retain MFI compatibility and introducing a spiking patch encoding layer to enhance local feature extraction capabilities. As a result, we establish an efficient multi-stage spiking MLP network that blends effectively global receptive fields with local feature extraction for comprehensive spike-based computation. Without relying on pre-training or sophisticated SNN training techniques, our network secures a top-1 accuracy of 66.39\% on the ImageNet-1K dataset, surpassing the directly trained spiking ResNet-34 by 2.67\%. Furthermore, we curtail computational costs, model parameters, and simulation steps. An expanded version of our network compares with the performance of the spiking VGG-16 network with a 71.64\% top-1 accuracy, all while operating with a model capacity 2.1 times smaller. Our findings highlight the potential of our deep SNN architecture in effectively integrating global and local learning abilities. Interestingly, the trained receptive field in our network mirrors the activity patterns of cortical cells. Source codes are publicly accessible at \url{https://github.com/EMI-Group/mixer-snn}.

\end{abstract}


\IEEEpeerreviewmaketitle

\section{Introduction}

\IEEEPARstart{O}{ver} the years, Convolutional Neural Networks (CNNs) have garnered considerable success within the field of computer vision. However, the advent and subsequent accomplishments of Transformer \cite{vaswani2017attention} in natural language processing have led to the emergence of the Vision Transformer (ViT) \cite{DBLP:journals/corr/abs-2010-11929} as a promising contender. The ViT model replaces the convolution operation in CNNs with the self-attention operation from the Transformer, enabling it to model visual relationships across different spatial locations of an image. ViT and its later versions \cite{yuan2021tokens, wang2021pyramid} have showcased performance on par with, or even superior to, CNNs. Unlike CNNs, which require intricate convolution kernel designs, ViTs utilize several standard Transformer blocks, thus minimizing manual manipulations and reduce inductive biases.

More recently, the MLP-Mixer \cite{tolstikhin2021mlp} researchers proposed a simpler approach, leveraging Multilayer Perceptrons (MLP) to further mitigate induction bias. The fundamental block of MLP-Mixer consists of two components: the channel-mixing MLP and the token-mixing MLP. The channel-mixing module performs MLP calculations on the channel dimension of the feature map, facilitating information exchange across different channels. Simultaneously, the token-mixing module conducts MLP calculations across varying positions of the feature map, enabling communication over the spatial dimension. By harmonizing these two modules, the MLP-Mixer can efficiently extract and integrate features from both dimensions, enhancing the model's overall performance. This methodology presents a simpler approach to reduce induction bias in comparison with previous models.

Rooted in computational neuroscience, SNNs have been extensively utilized for the modeling of brain dynamics and functions \cite{izhikevich2004model, brette2005adaptive, deco2008dynamic, gerstner2014neuronal, korcsak2022cortical}. Recently, the successes and challenges experienced by deep Artificial Neural Networks (ANNs) in tackling machine learning problems have fostered a growing curiosity towards SNNs. Researchers are exploring them as alternatives and are keen to leverage their bio-inspired properties to address similar issues.

The competence of SNNs has been convincingly demonstrated, with their performance showing promising results
\cite{leng2016spiking, leng2018spiking, wu2019direct, leng2020solving, li2021differentiable, fang2021deep, deng2022temporal, che2022differentiable,  zhang2023accurate}. This has been achieved through the astute application of adaptive learning algorithms \cite{bohte2000spikeprop, wu2018spatio, neftci2019surrogate, bellec2020solution} and the integration of efficient architectures borrowed from ANNs. These architectures span from the Boltzmann machines \cite{ackley1985learning}, prominent during the nascent stages of deep learning, to the currently dominant CNNs \cite{lecun1989backpropagation}.
Furthermore, to identify task-specific cells or network structures for SNNs, recent research has employed Neural Architecture Search (NAS) \cite{na2022autosnn, kim2022neural}. This approach not only strengthens the performance of SNNs but also broadens their potential application scope within the realm of machine learning.

While these networks have achieved significant reductions in spike occurrence and established new benchmarks in image classification, their underlying architectures still closely resemble deep CNNs. Recently, ANNs equipped with visual attention and transformer mechanisms \cite{DBLP:journals/corr/abs-2010-11929, liu2021swin} have outperformed pure CNNs by learning global image dependencies. However, these mechanisms typically rely on matrix multiplication and softmax functions, presenting a contradiction to the Multiplication-Free Inference (MFI) principle that characterizes SNNs \cite{roy2019towards, rathi2021diet}. Recent research, such as the SNN-MLP study \cite{li2022brain}, offers a hybrid model that blends MLP and SNN neurons, effectively enhancing the MLP model's performance. Nonetheless, this model does not conform to a fully spiking architecture, suggesting that it does not constitute a strict SNN in the traditional sense.

In this study, we propose a novel approach to implementing MLPs to be inherently compatible with the  MFI principle in a spike-based format. This partially comes on the heels of recent findings illustrating the equal efficiency of MLPs and transformers \cite{tolstikhin2021mlp}. However, challenges persist as the original MLP-Mixer architecture for ANNs involves real-valued matrix multiplication, thus violating the MFI principle.
To overcome these challenges, our contribution is twofold. First, we design a spiking MLP-Mixer architecture by utilizing MFI-friendly batch normalization (BN) combined with lightweight axial sampling in the token block. This architecture enables us to create a multi-stage spiking-MLP network, facilitating full spike-based computation. Second, we propose a spiking patch encoding module, anchored on a directed acyclic graph structure. This module replaces the original patch partition for downsampling, thereby enhancing local feature extraction capabilities of the MLP network. Moreover, our study underscores the critical role of skip connection configuration in achieving an optimal spiking MLP-Mixer design.
Our efforts yield the following primary outcomes:
\begin{itemize}
\item We successfully engineer an efficient spiking MLP-Mixer using MFI-friendly BN and lightweight axial sampling in the token block, further emphasizing the importance of optimal skip connection configuration.
\item By proposing a spiking patch encoding module, we enrich local feature extraction and enable downsampling, allowing for a multi-stage spiking-MLP network fully operating on spike-based computation.
\item Our network delivers a 66.39\% top-1 accuracy on the ImageNet-1K classification, marking a 2.67\% improvement over the current state-of-the-art deep spiking ResNet-34 network. Moreover, the network operates with a similar model capacity, but with 2/3 of its simulation steps and significantly reduced computation cost. A larger variant of our network achieves a 71.64\% top-1 accuracy, representing a 7.92\% improvement and rivalling the spiking VGG-16 network. All of this is achieved without resorting to advanced training techniques.
\item When fine-tuned on CIFAR10, CIFAR100, and CIFAR10-DVS datasets, our pre-trained networks on ImageNet set new benchmarks for SNNs, registering accuracies of 96.08\%, 80.57\%, and 81.12\% respectively. This showcases the broad applicability of our architectural design as pre-trained models.
\end{itemize}

The rest of this paper is organized as follows: Section II delves into the related works MLPs and SNNs, two distinct deep learning approaches; Section III first provides an overview of our network architecture, followed by a detailed discussion on the vital Spiking MLP Block module; Section IV describes the tests conducted on three distinct datasets (ImageNet-1K, CIFAR10/100, and CIFAR10-DVS) for classification tasks, and presents our findings, including results from ablation, network spiking rate, and neuronal weight visualization experiments; finally, Section V encapsulates the results and explores the implications of our work.

\section{Related Work\label{sec:related}}

\subsection{Multi-Layer Perceptrons in Deep Learning}
The recent advancements in attention and transformer mechanisms, particularly in speech \cite{vaswani2017attention} and vision tasks \cite{DBLP:journals/corr/abs-2010-11929}, have stimulated further investigations into similar mechanisms but in forms more biologically plausible. Studies outlined in \cite{widrich2020modern, ramsauer2020hopfield} demonstrated that the attention mechanism of transformers is equivalent to the update rule of modern Hopfield networks with continuous states. The high storage capacity of these networks in large-scale multiple instance learning was also highlighted. However, akin to the original transformer, these networks involve matrix multiplication and softmax function, both of which are incompatible with spike-based computation.

In the quest for an alternative architecture, the ground-breaking work of the MLP-Mixer \cite{tolstikhin2021mlp} offered a model relying solely on MPLs, eschewing both convolution and self-attention layers. The MLP-Mixer consists of two unique parts - the channel-mixing MLP and the token-mixing MLP. The channel mixing module primarily focuses on conducting MLP computations within the channel dimension of the feature map, leading to an exchange of information among various channels and thereby fostering the extraction of effective features. Conversely, the token mixing module carries out MLP computations among different feature map positions. This enhances communication across spatial dimensions, fostering integration of diverse features from various map locations. Together, these modules equip the MLP-Mixer model with a robust proficiency for feature extraction and integration from both dimensions.

The MLP-Mixer model offers a simple yet powerful method to mitigate induction bias, surpassing previous models. This straightforward approach has proven highly successful in boosting model performance and has drawn considerable attention from the machine learning community.

Freed from the inductive biases of local connectivity and self-attention, the token-mixing MLP features enhanced flexibility and superior fitting capability \cite{zhao2021battle, liu2022we}. However, it also displays a higher susceptibility to overfitting and typically relies on pre-trained models with large-scale datasets to achieve competitive performances. These performances are on par with image classification benchmarks such as Vision Transformer (ViT) and Convolutional Neural Networks (CNNs).
To overcome the issues of over-parametrization and overfitting associated with MLP-Mixer, a recent work proposed the SparseMLP \cite{tang2022sparse}. This approach adopts a multi-stage pyramid network structure and applies axial sampling, as opposed to full sampling, in the token mixing MLP. This concept has also been echoed in other MLP variants \cite{hou2022vision, tatsunami2021raftmlp, wang2022dynamixer}.

\subsection{Spiking Neural Networks in Deep Learning}
Spiking Neural Networks (SNNs) have established their prowess in effective information processing, primarily due to their bio-inspired mechanisms \cite{wysoski2010evolving, tavanaei2019deep, han2020deep, auge2021survey}. Deriving their roots from computational neuroscience, SNNs are frequently employed to model brain dynamics and functions \cite{izhikevich2004model, brette2005adaptive, deco2008dynamic, gerstner2014neuronal, korcsak2022cortical}. As machine learning evolves through its successes and challenges, SNNs have emerged as promising alternatives with researchers exploring their biological properties for functional advantages in similar applications.

Although SNNs were not originally designed for gradient-based supervised learning, their bio-inspired architecture and capability to process spiking neural activity have led to a surge in popularity. Traditional training methods for SNNs, such as Spike Timing-Dependent Plasticity (STDP) \cite{masquelier2007unsupervised}, while effective, pose limitations in incorporating global information. This shortfall can slow convergence speed and impede applicability to large models. 

Consequently, there has been a burgeoning interest in alternative SNN training techniques such as Backpropagation Through Time (BPTT) and surrogate gradients, which promise more efficient and effective learning for large-scale neural networks. Wu \textit{et al.} advocated explicitly iterative LIF neuron training to enhance speed and accuracy \cite{wu2019direct}, while Zheng \textit{et al.} proposed a threshold-dependent batch normalization to streamline the training process \cite{zheng2021going}. Despite the advantages of gradient-based training, such as the production of efficient SNNs requiring minimal time steps, SNNs lag behind CNNs trained similarly, in terms of accuracy. This discrepancy stems from the complex spiking dynamics of SNNs, creating unique challenges in accurately modeling the underlying computations and optimizing network performance.

A different approach to creating SNN models involves transforming pre-trained ANN/CNN models into spiking neural networks. This strategy retains the original models' accuracy by first training non-spiking ANNs/CNNs using standard methods, followed by a conversion into spiking neural networks \cite{rueckauer2017conversion}. Despite the ease of this conversion process, it poses significant challenges, primarily requiring larger time steps to compensate for the reduced accuracy resulting from the transition from full-precision to binary output. Consequently, recent research has focused on integrating the conversion and training processes, exemplified by novel methodologies such as progressive conversion \cite{severa2019training} and conversion during initialization \cite{rathi2020enabling}.

Recently, the exploration of Neural Architecture Search (NAS) has led to the discovery of task-specific cells or network structures applicable to SNNs \cite{na2022autosnn, kim2022neural}. These networks have achieved spike reduction and set new benchmarks in image classification. While these SNNs resemble deep CNNs, significant differences have emerged, such as alterations to traditional convolutional layers. In contrast, artificial neural networks that use visual attention and transformer mechanisms have recently outperformed pure CNNs in capturing global image dependencies \cite{DBLP:journals/corr/abs-2010-11929, liu2021swin}. However, these models typically require heavy matrix multiplication and normalization operations, which contravene the Multiplication-Free Inference (MFI) principle integral to SNNs \cite{roy2019towards, rathi2021diet}. Notably, these operations could lead to increased power consumption on neuromorphic hardware, which often possesses limited computational resources \cite{horowitz20141, davies2021advancing, merolla2014million, furber2014spinnaker, kungl2019accelerated}.

\subsection{Discussion}
The structural simplicity and efficacy of MLPs hint at a promising network paradigm. However, previous implementations of MLPs with non-spiking full-precision neurons fail to fulfill the MFI principle and eschew spike-based communication, rendering the network non-identical to typical SNNs \cite{li2022brain}. Therefore, a pressing challenge lies in designing an MLP-based architecture compatible with spike-based computation and aligned with the MFI principle.

One promising direction is to combine the advantages of MLPs and SNNs, a strategy that could bridge the performance gap between spike-based models and state-of-the-art deep learning models. It would also potentially unlock the advantages of spiking computation, such as low energy usage and robustness to noise. However, to realize this promise, researchers must overcome several hurdles, including addressing the challenge of matrix multiplication and softmax operations, as these operations are incompatible with the spike-based computation and the MFI principle inherent to SNNs.

Overall, the emergence and development of SNNs have shown promising potential for efficient and biologically inspired computation. As the research progresses, SNNs could pave the way for a new generation of neural networks that bridge the gap between artificial and biological systems, ultimately advancing our understanding of both machine learning and neurobiology.

\section{Proposed Approach\label{sec:approach}}

This section introduces our novel approach to addressing the challenges associated with the implementation of the MLP-Mixer in spiking neural networks. We propose a new network architecture, an optimized version of the MLP-Mixer that leverages the strengths of SNNs, and a unique implementation of the Leaky Integrate-and-Fire (LIF) neuron model. Key elements include a multi-stage pyramid network structure, the Spiking Patch Encoding (SPE) module, and the Spiking MLP-Mixer. We then elaborate on the structure of the spiking token block and the spiking channel block respectively, and discuss the implementation of the LIF spiking neuron model within the network.

\subsection{Network Architecture}
\begin{figure}[h!]
  \centering
  \includegraphics[width=0.7\linewidth]{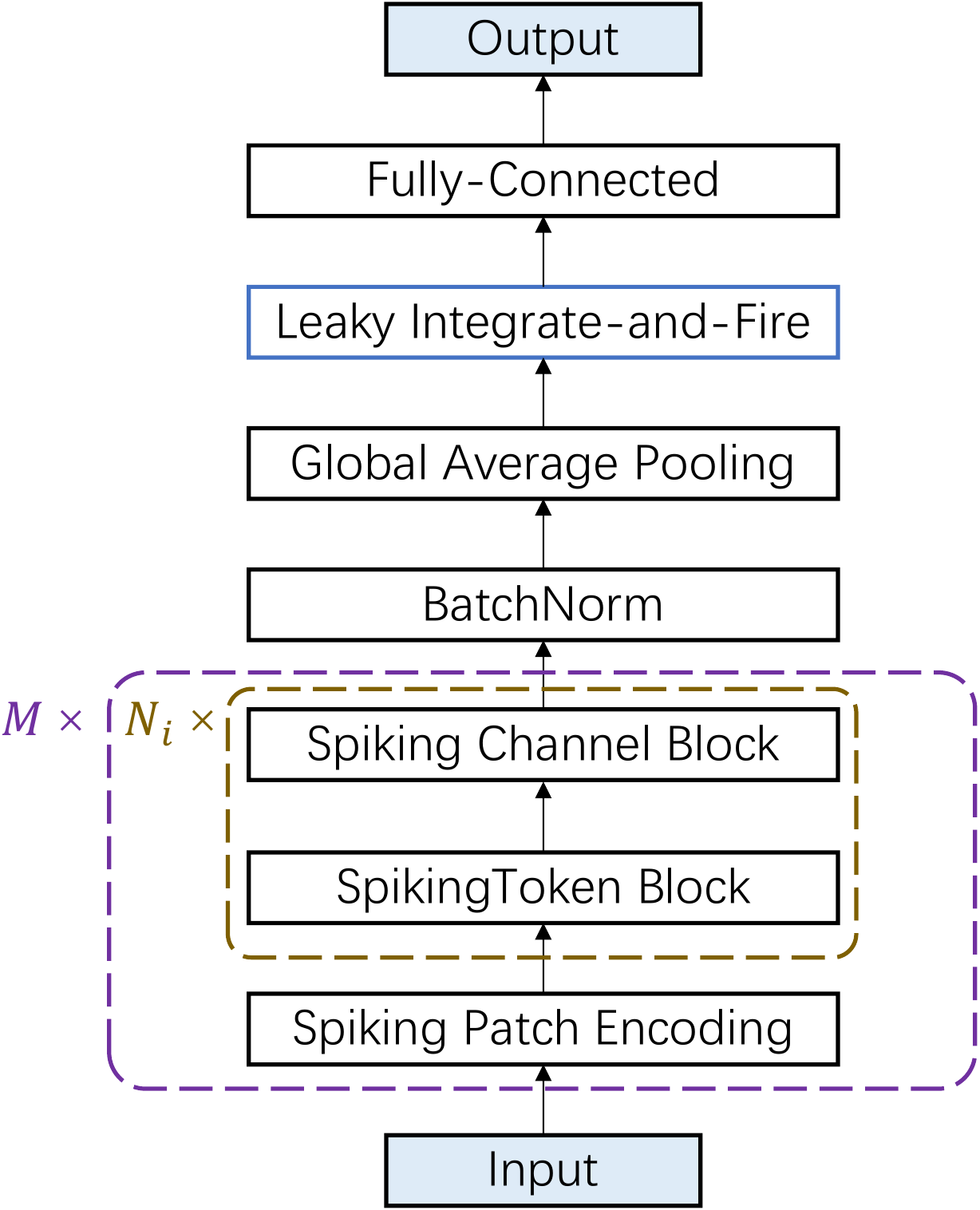}
  \caption{The overall network architecture. The multi-stage network is downsampled with an SPE module at each stage. Within each stage, the SPE is followed by a sequence of spiking MLP-Mixers with identical architecture, each containing a spiking token block with axial sampling and a spiking channel block with full sampling.}
  \label{fig1:snn_mlp}
\end{figure}

The architecture used by MLP-Mixer \cite{tolstikhin2021mlp} is characterized by an \emph{isotropic} design. This design holds the input and output resolutions constant across different layers. This type of structure tends to create a profuse number of parameters, leading to overfitting when training on medium-scale datasets, such as ImageNet-1K. To overcome this issue, we utilize a multi-stage pyramid network architecture as suggested by \cite{liu2021swin, chu2021twins, tang2022sparse}. Fig. \ref{fig1:snn_mlp} provides a visual representation of our network architecture.

\begin{figure*}[t!]
\centering
\begin{subfigure}{0.3\textwidth}
    \centering
      \includegraphics[width=0.9\linewidth]{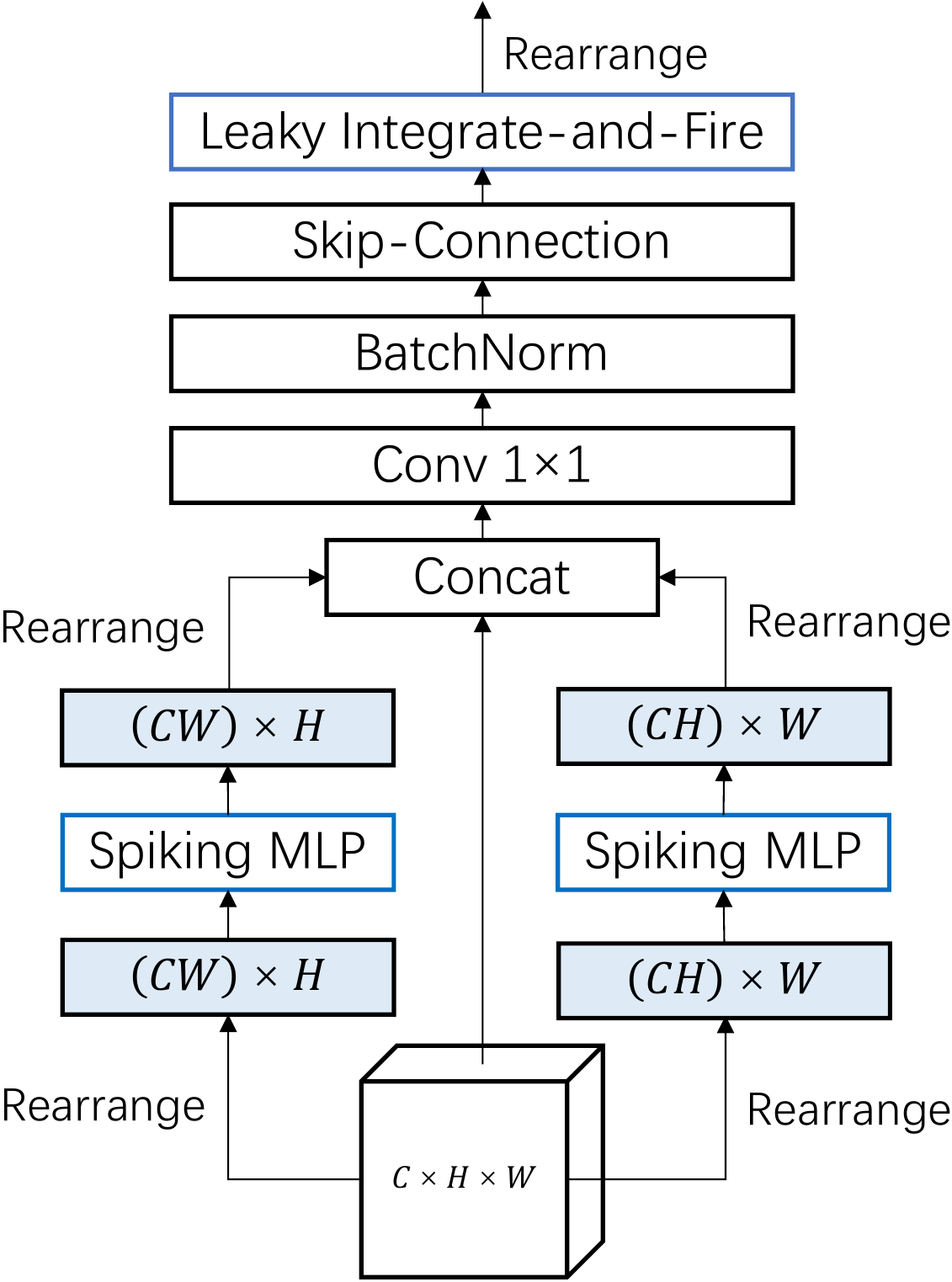}
        \caption{Spiking Token Block.}
\end{subfigure}
\begin{subfigure}{0.3\textwidth}
    \centering
      \includegraphics[width=0.6\linewidth]{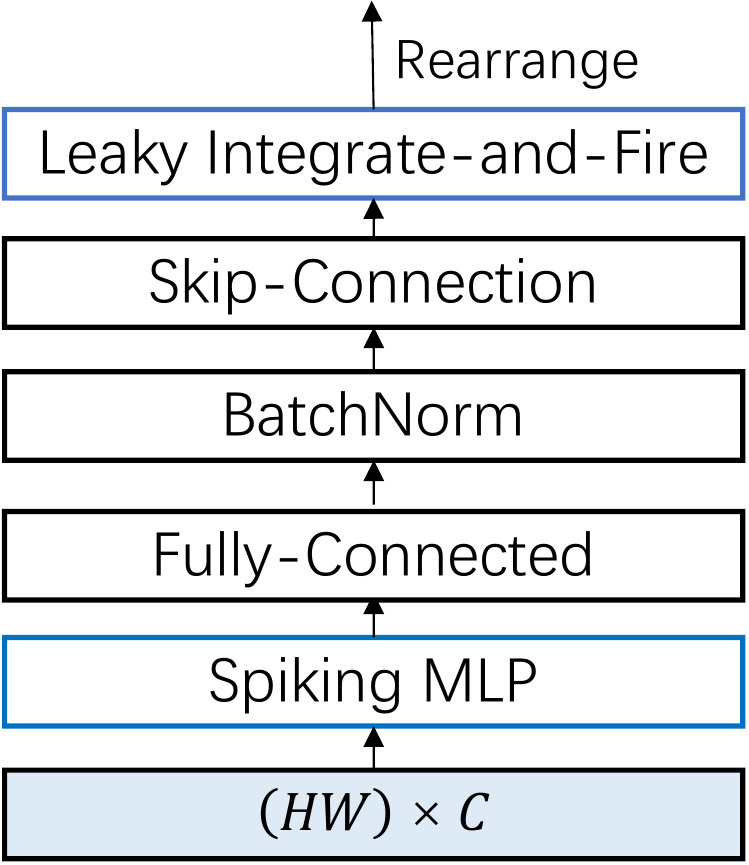}
      \vspace*{10mm}
        \caption{Spiking Channel Block.}
\end{subfigure}
\begin{subfigure}{0.3\textwidth}
    \centering
      \includegraphics[width=0.6\linewidth]{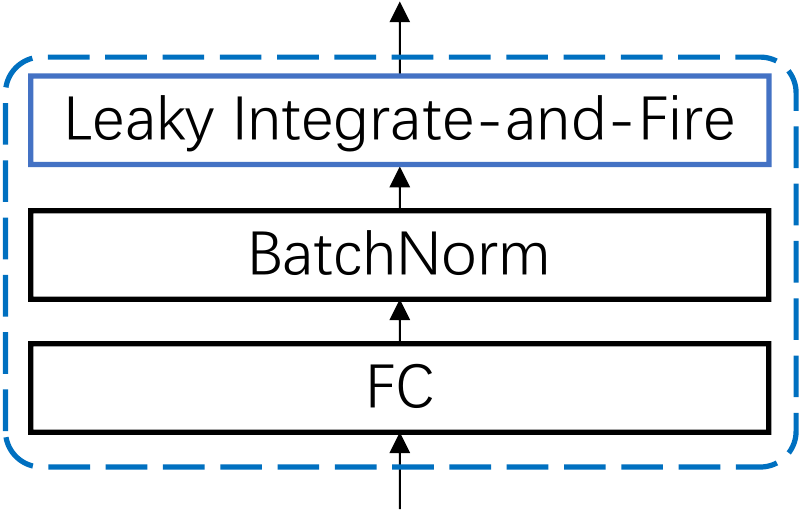}
            \vspace*{15mm}
        \caption{Spiking MLP Block.}
\end{subfigure}
  \caption{Key blocks in the proposed network architecture.}
    \label{fig:architecture}
\end{figure*}
\begin{figure*}[h!]
\centering
\begin{subfigure}{0.7\textwidth}
    \centering
  \includegraphics[width=\linewidth]{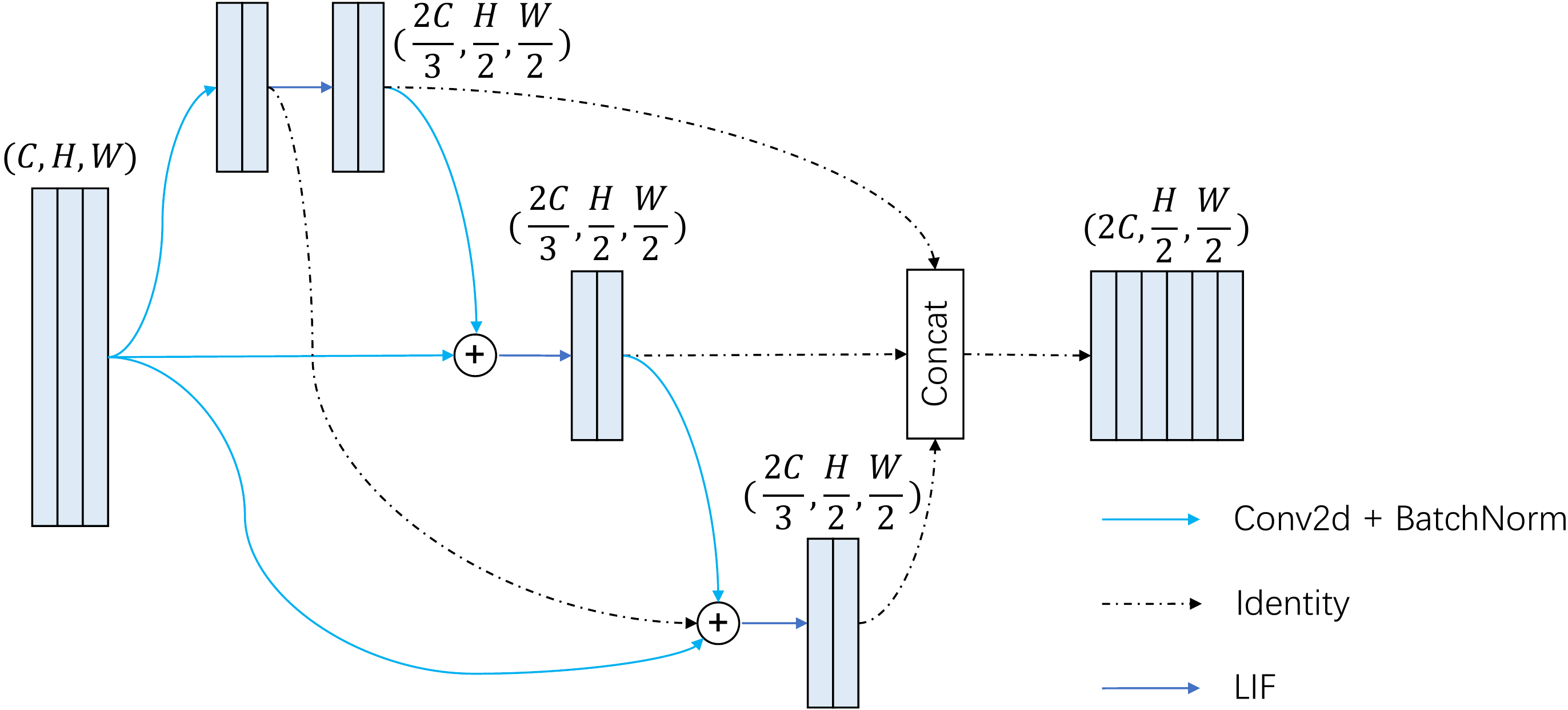}
\end{subfigure}
  \caption{Spiking patch encoding with a directed acyclic graph structure. The structure adheres to the MFI principle, with additions performed on BN states and multiplications performed between convolution weights and binary spikes.}
  \label{fig2:snn_cell}
\end{figure*}

Specifically, our approach starts with an RGB input image of size $3\times$H$\times$W. The original MLP-Mixer breaks this image down into non-overlapping patches of size $p \times p$, subsequently projecting the new channel dimension of $3 \times p^2$ to a hidden dimension of $C_1$. This process parallels a 2D convolution operation with a kernel and stride size of $p$, and an output dimension of $C_1$. Although the MLP-Mixer benefits from a global receptive field, it lacks the necessary inductive bias for learning local features. To address this deficiency, we develop an SPE module to replace the original patch partitioning approach. We construct the SPE with a spike-based directed acyclic graph, following the design outlined in a recent work \cite{che2022differentiable}, as illustrated in Figure \ref{fig2:snn_cell}.
Since this structure draws inspiration from DARTS \cite{liu2018darts}, we term it as a spiking cell, which comprises three nodes. Each node receives the same input from the previous stage, which is processed through a convolution operation followed by Batch Normalization (BN). Node pairs (1,2) and (2,3) are interconnected through a convolution operation followed by BN, whereas node pair (1,3) is connected via an identity connection. Within each node, multiple operations are added after BN, with the subsequent spiking activation serving as the output of the node. The cell's output is the concatenated spiking states of all nodes.

The output of the spiking cell is then fed into a sequential of spiking MLP-Mixers which finally output a spiking feature map of the same shape. 
In the following stages, this process is repeated with the input feature consecutively downsampled in spatial dimension and expanded in channel dimension, both with a ratio of 2.
The output of the final stage is fed into a spiking classification head with a spiking MLP block consisting of a fully connected (FC) layer followed by BN and a spiking activation, with the FC layer performing full sampling on the flattened spatial dimension of the feature map.
Finally, a linear classifier projects the output of spiking MLP block to a label layer. The loss function is a cross entropy function with the network output averaged over simulation time steps.

\subsection{Spiking MLP-Mixer}
A straightforward adoption of existing MLP block designs from artificial neural networks (ANNs) by substituting the real-valued activation function with a spiking activation function (SAF) results in a violation of the MFI principle of spike-based computation. In both the token and channel blocks of the original MLP-Mixer, the fully connected (FC) operation is carried out on the layer normalization (LN) \cite{ba2016layer} state of the feature map, leading to real-valued matrix multiplication. To rectify this, we move the normalization process to after the FC operation and replace LN with batch normalization (BN). This adjustment enables the parameters of the latter to be integrated with the linear projection weights during inference, aligning with the MFI principle \cite{ioffe2015batch, rueckauer2017conversion}.

The spiking MLP-Mixer comprises a token block and a channel block, each of which includes multiple FC layers with spiking activation, as depicted in Fig.~\ref{fig:architecture}.

\subsubsection{Spiking Token Block}
Given input patched image $\displaystyle \mI$ of shape $C \times H \times W$,  the token block of the original MLP-Mixer flattens its spatial dimension to form a 2D tensor. The process performs full sampling on the token dimension with weights shared across the channel dimension, thus creating a global receptive field that can be heavily parameterized for small patch sizes.
Drawing inspiration from previous works \cite{tatsunami2021raftmlp, hou2022vision, tang2022sparse, wang2022dynamixer}, we design a two-branch structure for the spiking token block. This structure separately encodes the feature representation along the horizontal and vertical spatial dimensions. It learns long-range dependencies along one direction while preserving positional information along the other.

Given input patched image $\displaystyle \mI$ of shape $C \times H \times W$,  
the token block of the original MLP-Mixer flattens its spatial dimension to form a 2D tensor. The process performs full sampling on the token dimension with weights shared across the channel dimension, thus creating a global receptive field that can be heavily parameterized for small patch sizes. Drawing inspiration from previous works \cite{tatsunami2021raftmlp, hou2022vision, tang2022sparse, wang2022dynamixer}, we design a two-branch structure for the spiking token block. This structure separately encodes the feature representation along the horizontal and vertical spatial dimensions. It learns long-range dependencies along one direction while preserving positional information along the other. As illustrated in Figure \ref{fig1:snn_mlp}(c), each branch comprises an FC layer followed by a BN and an SAF. The two binary feature maps are then concatenated along the channel dimension, along with an identity connection from the input feature map, and projected back to $C$ through an FC layer with BN and an SAF. Furthermore, we add a skip connection from the BN states of the SPE.

In a single stage of the Spiking MLP-Mixer, assuming the input $\mX$ of shape $C \times (HW)$ and the output $\mY$, we can express the Spiking Token Block as follows:
\begin{align}
\displaystyle \mI &= f(\displaystyle \mX), \\
\displaystyle \mU_{h} &= \mathrm{BN}(\displaystyle \mW_h \displaystyle \mI^h), \\
\displaystyle \mU_w &= \mathrm{BN}(\displaystyle \mW_w \displaystyle \mI^w), \\
\displaystyle \mY &= \mathrm{BN} (\displaystyle \mW_f \mathrm{Concat}[f(\displaystyle \tilde{\mU}_h),f(\displaystyle \tilde{\mU}_w), \displaystyle \mI])+\displaystyle \mX,
\end{align}
where $\mathrm{BN}$ signifies batch normalization, $f$ represents an SAF, $\displaystyle \mW_h$ and $\displaystyle \mW_w$ are the token FC weights acting on the height and width dimensions with shapes $H \times H$ and $W \times W$ respectively. $\displaystyle \mW_f$ are the branch fusion weights with a shape of $C \times 3C$. Superscripts $h$ and $w$ denote the matrix reshape operation that retains the height, width, and channel dimensions, and $\tilde{}$ denotes the reshape operation that flattens the spatial dimension with $\displaystyle \tilde{\mU}_h$ and $\displaystyle \tilde{\mU}_w$ of shape $C\times (HW)$.

\subsubsection{Spiking Channel Block}
The channel block, comprised of two spiking FC layers, takes the transposed output from the token block as its input. The initial layer expands the channel dimension by a predetermined ratio, $\alpha$, and the subsequent layer restores this dimension to its original state. A skip connection is integrated between the BN state of the second layer and the aggregated BN state from the token block.

In a single stage of the Spiking MLP-Mixer, given the input $\mX$ with shape $C \times (HW)$ and the output $\mY$, we can formulate the Spiking Channel Block as follows:
\begin{align}
\displaystyle \mI &= f(\displaystyle \mX), \\
\displaystyle \mV &= \displaystyle \mX^c + \mathrm{BN}(\displaystyle \mW_{c2} f({\mathrm{BN}(\displaystyle \mW_{c1} \displaystyle \mI^c)})), \\
\displaystyle \mY &= f(\displaystyle \tilde{\mV}),
\end{align}
where $\displaystyle \mW_{c1}$ and $\displaystyle \mW_{c2}$ are the channel FC weights with shapes $\alpha C \times C$ and $C \times \alpha C$ respectively, and $\alpha$ is the channel expansion ratio.

To counteract the gradient vanishing issue commonly observed in deep SNNs, we incorporate skip connections between the final BN states of both the token and channel blocks, as well as the BN states from the SPE. Our ablation studies highlight the significance of this design in achieving optimal network efficiency. 

It is noteworthy that all matrix multiplications involve a real-valued matrix and a binary matrix, which implies that the spiking MLP-Mixer essentially operates without requiring any multiplications. In subsequent spiking MLP-Mixers, the input tensor is the output generated from the channel block of the preceding Mixer.

\subsection{LIF Spiking Neuron}
In our study, we employ the Leaky Integrate-and-Fire (LIF) neuron model, which is characterized by a hard threshold and decay input, as described by the following equation:
\begin{align}
    \displaystyle \vu^{t, \mathrm{pre}} &= \displaystyle \vu^{t-1} + \frac{{\displaystyle \vz}^{t} - \displaystyle \vu^{t-1}}{\tau}, \\
\displaystyle \vy^{t} &= g(\displaystyle \vu^{t, \mathrm{pre}}), \\
\displaystyle \vu^{t} &= (1 - \displaystyle \vy^{t})\displaystyle \vu^{t, \mathrm{pre}}, 
\label{eq1:LIF}
\end{align}
where $t$ signifies the time step, $\tau$ represents the membrane time constant, and $\vu$ is the membrane potential, denoted by a bold, italicized letter to represent a vector. $\vy$ stands for the spike output and $g$ is a threshold function. $\vz^{t} = \mW \vy^{t, \mathrm{pre}}$ refers to the synaptic input, where $\mW$ is the weight matrix, and $\vy^{t, \mathrm{pre}}$ signifies the afferent spikes originating from pre-synaptic neurons.

When the membrane potential of a neuron surpasses a certain threshold $V_{th}$, the neuron fires a spike and conveys it to the post-synaptic neuron. This action is followed by a hard reset of the membrane potential. If the membrane potential does not exceed the threshold, the neuron does not transmit any signals, as delineated by:
\begin{equation}
y^{t}_i =\left\{
\begin{aligned}
\,\,\, & 1 \,\,\,\,\,  \text{ if } u^{t, \mathrm{pre}}_i \geq V_{th} \\
\,\,\, & 0 \,\,\,\,\,  \text{ otherwise }
\end{aligned}
\right. .
\end{equation}
Previous works have shown that training thresholds or membrane dynamics could potentially improve network performances \cite{rathi2021diet, fang2021incorporating, zhang2022discrete}. Since this work focuses more on architectures, we set $\tau=2$ and $V_{th}=1$ during experiments. Given a loss $L$ and using the chain rule, the weight update of a Spiking Neural Network (SNN) can be expressed as follows:
\begin{equation}
    \frac{\partial L}{\partial \displaystyle \mW}= \sum_t\frac{\partial L}{\partial \displaystyle \vy^{t}} \frac{\partial \displaystyle \vy^{t}}{\partial \displaystyle \vu^{t, \mathrm{pre}}} \frac{\partial \displaystyle \vu^{t, \mathrm{pre}}} {\partial \displaystyle \vz^{t}} \frac {\partial \displaystyle \vz^{t}} {\partial \displaystyle \mW},
    \label{eq:stbp}
\end{equation}
where $\frac{\partial \vy^{t}}{\partial \vu^{t, \mathrm{pre}}}$ is the gradient of the firing function, which equals zero at all points except at the threshold. The surrogate gradient method leverages continuous functions to estimate the gradients. Various continuous functions have been used in previous studies, such as rectangular \cite{zheng2021going}, triangular \cite{bellec2018long}, and exponential curves \cite{shrestha2018slayer, zenke2018superspike}, among others. In our experiments, we chose to use the sigmoid function.

\section{Experiments\label{sec:exper}}
In order to thoroughly evaluate the performance and capabilities of our model, a variety of image classification benchmarks are utilized. These include the ImageNet-1K \cite{krizhevsky2017imagenet}, CIFAR10/100 \cite{cifar}, and CIFAR10-DVS \cite{li2017cifar10} datasets. The SpikingJelly \cite{SpikingJelly} toolkit, complemented by its cupy backend, enables rapid simulation of Leaky Integrate-and-Fire (LIF) neurons. 
The SPE module incorporated in our architecture utilizes $3\times3$ convolution operations. The stride for input-to-node connections is set to 2, while node-to-node connections have a stride of 1.
To avoid statistical bias, the batch mean and variance are synchronized across each device, following the approach suggested by \cite{li2021differentiable}.

The following sections will present results from the aforementioned datasets, discuss an ablation study, assess the network spiking rate and computational cost, and finally visualize our findings.

\subsection{Results on ImageNet-1K}
The ImageNet-1K dataset is comprised of a training set, validation set, and a test set, with 1.28M, 50K, and 100K 224$\times$224 images, respectively. Our training settings are predominantly derived from \cite{tang2022sparse}, encompassing data augmentation. We adopt a cosine decay learning rate strategy, initializing the learning rate at 0.1 and gradually lowering it to zero over 100 epochs. Stochastic Gradient Descent (SGD) is utilized as the optimizer with a momentum of 0.9.

In the first stage of the spiking MLP network, we use the original patch partitioning approach with a patch size of 4, while employing the SPE module for downsampling in subsequent stages. Three variants of the spiking MLP network are developed, specifically spiking MLP-SPE-T/S/B. The respective architectures are: {spiking MLP-SPE-T}: $C_1 = 78$, number of layers in each stage = \{2; 8; 14; 2\}; {spiking MLP-SPE-S}: $C_1 = 96$, number of layers in each stage = \{2; 8; 14; 2\}; {spiking MLP-SPE-B}: $C_1 = 108$, number of layers in each stage = \{2; 10; 24; 2\}. 

The expansion ratio of the channel FC layer is set to $\alpha =3$. To study the influence of the SPE module, we also train an MLP-S model with the original patch partition approach while other parts of the network remaining the same. The simulation step of SNN is set to $T=4$, 
where $T$ denotes the simulation step.
To better compare with other SNNs with larger time steps, we use time inheritance training (TIT) \cite{deng2022temporal} and obtain the $T=6$ result for MLP-SPE-T, where we initialize the network with the pre-trained MLP-SPE-T ($T=4$) model and fine-tune for 50 epochs with $T=6$ using a cosine learning rate decaying from 0.1 to 0.

Our model is compared to other state-of-the-art SNNs, including those using direct training and Artificial Neural Network (ANN)-to-SNN conversion methods. The results are presented in Table \ref{tab:imgt}, where we report the top-1 accuracy along with the number of model parameters and simulation steps.

\begin{table*}[h!]
\small
\centering
\caption{Results on ImageNet-1K (T denotes simulation length)}
\label{tab:imgt}

\begin{tabular}{ccrrl}
\toprule Method & Architecture & Model Size & T & Accuracy[$\%$] \\
\midrule 
ANN-SNN \cite{hu2018spiking}& ResNet-34 & 22M & 768 & $71.6$ \\
ANN-SNN \cite{sengupta2019going} & VGG-16 & 138M & 2500 & $69.96$ \\
S-ResNet \cite{9597475} & ResNet-50 & 26M & 350 & $73.77$ \\
Hybrid training \cite{rathi2020enabling} & ResNet-34 &22M & 250 & $61.48$ \\
Hybrid training \cite{lee2020enabling} & VGG-16 & 138M &  250 & $65.19$ \\
Tandem Learning \cite{9492305} & AlexNet & 62M & 10 & $50.22$ \\
STBP-tdBN \cite{zheng2021going} & ResNet-34 & 22M & 6 & $63.72$ \\
TET \cite{deng2022temporal} & ResNet-34 & 22M & 6 & $64.79$ \\
STBP-tdBN \cite{zheng2021going} & ResNet-34-large & 86M & 6 & $67.05$ \\
Diet-SNN \cite{rathi2021diet} & VGG-16 & 138M & 5 & $69.00$ \\
SpikeDHS \cite{che2022differentiable} & SpikeDHS-CLA-large & 58M & 6 & $67.96$ \\
\textbf{Spiking MLP} (our model) & MLP-SPE-T & \textbf{25M}  & \textbf{4} & $\mathbf{66.39}$ \\
\textbf{Spiking MLP} (our model) & MLP-SPE-T & \textbf{25M}  & \textbf{6} & $\mathbf{69.09}$ \\
\textbf{Spiking MLP} (our model) & MLP-S & \textbf{34M} & \textbf{4} & $\mathbf{63.25}$ \\
\textbf{Spiking MLP} (our model) & MLP-SPE-S & \textbf{38M}  & \textbf{4} & $\mathbf{68.84}$ \\
\textbf{Spiking MLP} (our model) & MLP-SPE-B & \textbf{66M}  & \textbf{6} & $\mathbf{71.64}$ \\
\midrule
SpikFormer (SNN) \cite{zhou2022spikformer} & Spikformer-8-512& 30M & 4 & $73.38$  \\
SparseMLP (ANN) \cite{tang2022sparse} & sMLPNet-T & 24M  & - & $81.9$ \\
\bottomrule
\end{tabular}
\end{table*}
It can be observed that ANN-SNN conversion methods can indeed achieve high accuracy, but they require exceedingly lengthy simulation steps. Hybrid training methods manage to reduce the simulation steps but remain considerably longer than methods using direct training. Among directly trained methods, and considering similar network capacity, MLP-SPE-T significantly outperforms STBP-tdBN ResNet-34 by 2.67\%, all the while requiring fewer simulation time steps. Post-TIT, the model achieves an accuracy of 69.09\% with $T=6$, exceeding ResNet-34-large, VGG-16 and the DARTs-based SpikeDHS models by 2.04\%, 0.09\%, 1.13\% respectively, while maintaining a model capacity more than 3, 5.5 and 2.3 times smaller. These results underscore the superior efficiency of our architecture.

The enhancement observed in MLP-SPE-S over MLP-S confirms that the SPE module can substantially improve network performance. This result attests to the efficacy of amalgamating global receptive fields with local feature extraction. However, it is important to note that our method is more accurately compared with the STBP-tdBN method since both use the same conventional temporal averaged loss function. The Temporal Error Transport (TET) method, which employs a temporal moment-wise loss function, has been demonstrated to outperform the temporal averaged method~\cite{deng2022temporal}. Note that the work of \cite{li2022brain} has applied temporal self-recurrent non-spiking neurons in MLP architecture, achieving an accuracy similar to transformer-based models on ImageNet-1K. However, due to the use of floating-point valued neurons, the model has no significant advantage in terms of energy cost compared with other ANNs. The SparseMLP \cite{tang2022sparse} in the ANN domain and SpikFormer \cite{zhou2022spikformer} using attention mechanism have significant higher accuracy compared with our model under similar parameter number. However, the sparse spiking activation of our architecture renders much lower energy cost, as demonstrated in Table \ref{tab:energy}.

\subsection{Results on CIFAR}
Both the CIFAR10 and CIFAR100 datasets consist of 50K training images and 10K testing images, each of size $32\times32$ pixels. For data augmentation, we implement random resized cropping and random horizontal flipping, in line with other studies. We follow a pre-training and fine-tuning paradigm, akin to other transformer and MLP works \cite{DBLP:journals/corr/abs-2010-11929, tolstikhin2021mlp}.

The pre-training phase employs the spiking MLP-SPE-T network trained on the ImageNet-1K dataset. We then reset the final classification output layer of the network and fine-tune it on the resized $224 \times 224$ CIFAR10/100 dataset. The fine-tuning phase uses a cosine decay learning rate strategy, starting from an initial value of 0.1 and gradually reducing to zero over 100 epochs. The optimizer we utilize is SGD with a momentum of 0.9.

Our results are benchmarked against other state-of-the-art SNNs, as depicted in Table \ref{tab:cifar}. Despite a larger network size, our model sets new records on both datasets, outperforming existing SNNs significantly. While comparing our results with directly trained SNNs might seem somewhat disproportionate, the outcomes convincingly illustrate the efficacy of the spiking MLP network as a pre-trained model applicable to other datasets. Note that our model also outperforms \cite{na2022autosnn}, a light weight SNN specifically optimized for balanced accuracy and energy cost, under much smaller timestep.

\begin{table*}[h!]
\small
\centering
\caption{Results on CIFAR (T denotes simulation length)}
\label{tab:cifar}

\begin{tabular}{cccrl}
\toprule Dataset & Method & Architecture & T & Accuracy[$\%$] \\
\midrule \multirow{6}{*}{CIFAR10} 
& TerMapping \cite{9328869} & VGG & 2800 & $93.75$ \\
& AugMapping \cite{9328869} & VGG & 300 & $93.90$ \\
& S-ResNet \cite{9597475} & ResNet-110 & 350 & $93.02$ \\
& Diet-SNN  \cite{rathi2021diet} & ResNet-20 & 10 & $92.54$ \\
& BRP-SNN \cite{9454259} & CNN & 20 & $57.08$ \\
& Tandem Learning \cite{9492305} & CifarNet & 8 & $90.98$ \\
& SAC \cite{10246307} & ResNet-18 & 6 & $93.74$ \\
& STBP-tdBN \cite{zheng2021going} & ResNet-19 & 6 & $93.16$ \\
& STBP-tdBN \cite{zheng2021going} & ResNet-19 & 4 & $92.92$ \\
& TET \cite{deng2022temporal} & ResNet-19 & 4 & $94.44 \pm 0.08$ \\
& AutoSNN \cite{na2022autosnn} & AutoSNN & 8 & $93.15$ \\
& SpikeDHS \cite{che2022differentiable} & SpikeDHS-CLA & 6 & $95.35$ \\
& \textbf{Spiking MLP} (our model) & MLP-SPE-T & $\mathbf{4}$ & $\mathbf{96.08}$ \\
\midrule 
\multirow{6}{*}{CIFAR100}
& S-ResNet \cite{9597475} & ResNet-110 & 350 & $70.62$ \\
& Diet-SNN \cite{rathi2021diet} & ResNet-20 & 5 & $64.07$ \\
& STBP-tdBN \cite{zheng2021going} & ResNet-19 & 6 & $71.12 \pm 0.57$ \\
& STBP-tdBN \cite{zheng2021going} & ResNet-19 & 4 & $70.86 \pm 0.22$ \\
& TET \cite{deng2022temporal} & ResNet-19 & 6 & $74.72 \pm 0.28$ \\
& TET \cite{deng2022temporal} & ResNet-19 & 4 & $74.47 \pm 0.15$ \\
& SpikeDHS \cite{che2022differentiable} & SpikeDHS-CLA & 6 & $76.15$ \\
& \textbf{Spiking MLP} (our model) & MLP-SPE-T & $\mathbf{4}$ & $\mathbf{80.57}$ \\
\bottomrule
\end{tabular}
\end{table*}

\subsection{Results on CIFAR10-DVS}
Neuromorphic datasets generally possess higher noise levels in comparison to static datasets, which heightens the risk of overfitting in well-optimized SNNs. Among all mainstream neuromorphic datasets, CIFAR10-DVS is recognized as one of the most challenging, presenting approximately 900 training samples for each label. Recent literature suggests an inclination towards complex architectures to handle this dataset effectively, albeit this strategy carries an inherent risk of overfitting without correspondingly enhancing the model's accuracy. Therefore, a pressing need exists for novel methodologies that can reduce the influence of noise and tackle the difficulties inherent in working with CIFAR10-DVS.

Similar to the CIFAR10/100 datasets, we employ the spiking MLP-SPE-T network, pre-trained on the ImageNet-1K dataset, as the pre-trained model for the CIFAR10-DVS dataset. We reset the final classification output layer of the network, and subsequently fine-tune it on the resized $224 \times 224$ CIFAR10-DVS dataset. Our fine-tuning strategy employs a cosine decay learning rate, starting with an initial value of 0.1 and progressively reducing to zero over 100 epochs. SGD, with a momentum of 0.9, is used as the optimizer.

We compare our results with those achieved by other state-of-the-art SNNs, as shown in Table \ref{tab:cifardvs}. Our experiments indicate that the spiking MLP-SPE-T outperforms the spiking ResNet-19 and even exceeds the performance of SNNs trained using Dspike.

\begin{table*}[h!]
\small
\centering
\caption{Comparison on CIFAR10-DVS (T denotes simulation length)}
\label{tab:cifardvs}
\begin{tabular}{ccrrl}
\toprule Method & Architecture & T & Accuracy[$\%$] \\
\midrule 
Tandem Learning \cite{9492305} & CNN & 10 & $65.59$ \\
STBP-tdBN \cite{zheng2021going} & ResNet-19 & 10 & $67.80$ \\
TET \cite{deng2022temporal} & VGG & 10 & $83.17$ \\
Dspike \cite{li2021differentiable} & ResNet-18 &  10 & $75.40$ \\
\textbf{Spiking MLP} (our model) & MLP-SPE-T & \textbf{10} & $\mathbf{81.12}$ \\
\bottomrule
\end{tabular}
\end{table*}

\subsection{Ablation Study}
\subsubsection{Normalization Methods}
\begin{table}[h!]
\small
\centering
\caption{Comparisons of different normalization methods on ImageNet-1K (T denotes simulation length)}
\label{tab:normalization}
\begin{tabular}{ccrrl}
\toprule Architecture & Normalization & Model Size & T & Accuracy[$\%$] \\
\midrule 
MLP-SPE-T & Batch Norm & 25M & 4 & $66.39$ \\
MLP-SPE-T & Layer Norm & 25M  & 4 & $65.94$ \\
MLP-SPE-T & TEBN & 25M & 4 & $62.33$ \\
MLP-SPE-T & tdBN & 25M & 4 & $60.25$ \\
\bottomrule
\end{tabular}
\end{table}
We conduct additional experiments focused on replacing the BN operation within the MLP-Block by other normalization methods including the original LN \cite{ba2016layer} in the MLP-Mixer and two recent temporal normalization methods specifically proposed for SNNs, i.e., tdBN \cite{zheng2021going} and TEBN \cite{duan2022temporal}. These experiments are conducted on the ImageNet-1k dataset, maintaining consistency with the hyperparameters used in the ImageNet experiments. The results of these experiments are presented in Table \ref{tab:normalization}. 
Our findings indicate that BN suggests more beneficial for accuracy improvement compared to other normalization methods. Additionally, employing BN allows the parameters of the latter to integrate with the linear projection weights during inference, aligning with the MFI principle.
\subsubsection{Skip Connection}
\begin{figure}[h!]
  \centering
  \includegraphics[width=\linewidth]{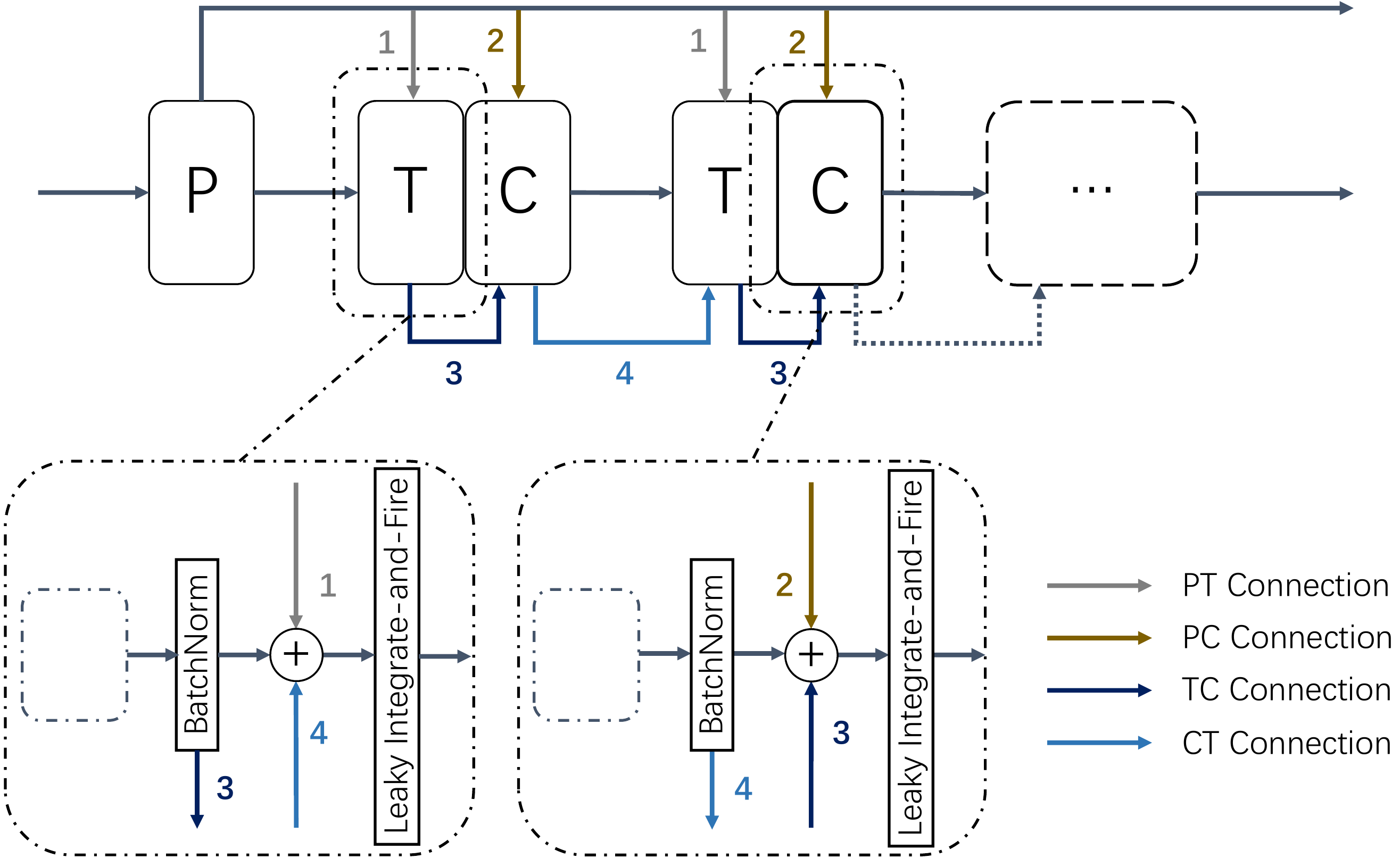}
  \caption{Potential skip connections for the spiking MLP-Mixer.}
  \label{fig:fig2}
\end{figure}

We further study the influence of different configurations of skip connections on the performance of the network. As shown in Figure \ref{fig:fig2}, we distinguish different skip connections, i.e patch block to token block (PT), patch block to channel block (PC), token block to channel block (TC) and channel block to the next token block (CT), with corresponding indices and colors. The skip connection from the patch block to the first token block is set as default. To shorten the simulation time, we set the initial channel number to 60 and directly train networks of different skip connection configurations from scratch on the CIFAR10 dataset for 100 epochs. The results are collected in Table \ref{tab:skip}. Networks without skip connections from the initial patch block to sequential token and channel blocks perform much worse than the others. This could be due to long range skip connections alleviating the gradient vanish problem of deep SNNs during training. Our spiking MLP-mixer adopts the optimal skip connection configuration of (PT, PC, TC).
\begin{table}
  \centering
\caption{Network performance under different skip connection configurations on CIFAR10}
\begin{tabular}{ccccc}
\toprule 
PT(1) & PC(2) & TC(3) & CT(4) & Accuracy[$\%$]\\
\midrule
\checkmark & \checkmark & & & $80.53$ \\
\checkmark & \checkmark & \checkmark &  & $81.35$ \\
\checkmark & \checkmark & \checkmark & \checkmark& $79.46$ \\
& & \checkmark & \checkmark & $10.53$ \\
& & & \checkmark & $10.00$ \\
\bottomrule
\label{tab:skip}
\end{tabular}
\end{table}

\subsection{Network Spiking Rate and Computational Cost}
\begin{figure*}[ht!]
\centering
\begin{subfigure}{0.9\textwidth}
    \centering
    \includegraphics[width=1\linewidth]{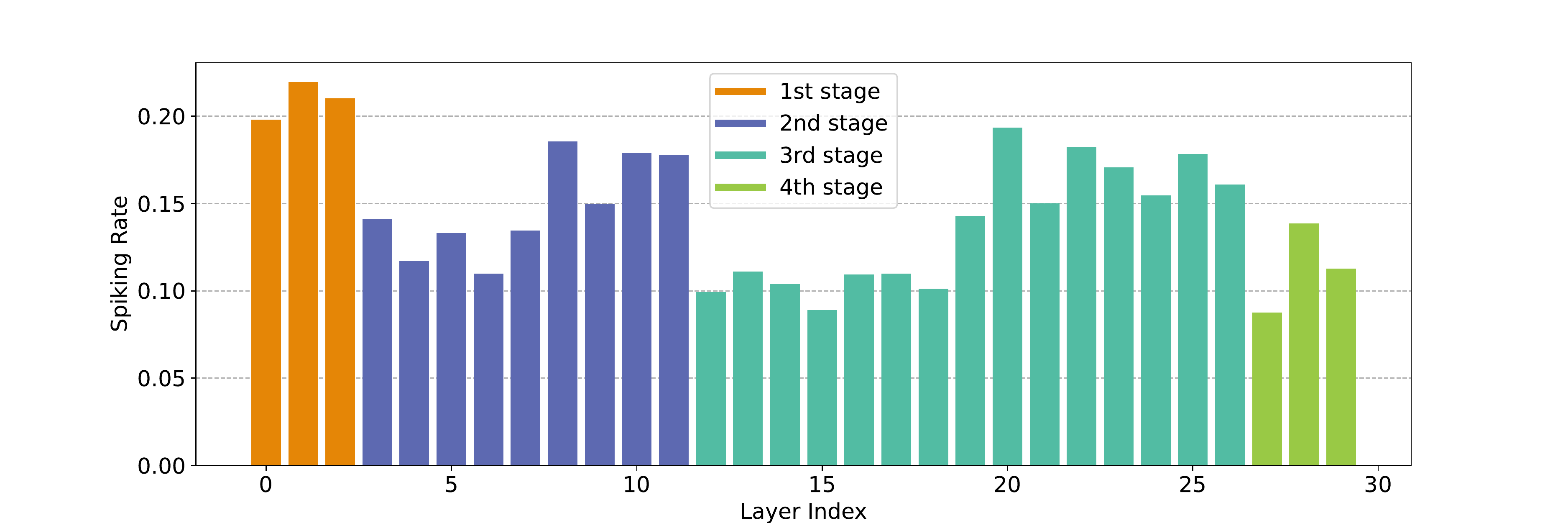}
\end{subfigure}
  \caption{Mean network spiking rate of MLP-SPE-T on the ImagNet-1K test set. We distinguish each stage with different colors.}
  \label{fig:fig3}
\end{figure*}
SNNs are recognized for their superior energy efficiency compared to dense-computing ANNs, a quality that primarily stems from the inherent event-driven and sparse computing characteristics of SNNs. Additionally, the MFI principle facilitates an entirely addition-based operation for SNNs, thus further diminishing their potential energy usage in contrast to ANNs that necessitate real-valued matrix multiplications. In this section, we assess the spiking rate and computation cost of the spiking MLP-SPE-T model, contrasting it with the spiking ResNet-34.

Figure \ref{fig:fig3} portrays the average spiking rate of each spiking MLP-Mixer on the ImageNet-1K test set. Our model demonstrates diverse sparse activity across different layers, yielding an average network spiking rate of 0.14. Following the methodologies established by \cite{li2021differentiable, rathi2021diet}, we compute the number of addition operations in the SNN using the formula $sTA$, where $s$ represents the average spiking rate (the total count of spikes divided by the total feature map size), $T$ indicates the simulation time step, and $A$ signifies the number of additions.

The findings are presented in Table \ref{comp}. The least amount of multiplication arises from the input layer, which receives floating-value images. To estimate the network's energy consumption, we refer to the study by \cite{horowitz20141} on the $45 \mathrm{nm}$ CMOS technology, a method also employed by previous works \cite{li2021differentiable, rathi2021diet}. According to this estimate, a single addition operation in an SNN costs $0.9\mathrm{pJ}$, whereas a multiply-accumulate (MAC) operation in ANNs consumes $4.6\mathrm{pJ}$. Notwithstanding their comparable model capacities, our network's computation cost is nearly half of that of the spiking ResNet-34. The SparseMLP \cite{tang2022sparse} in the ANN domain and the SpikFormer \cite{zhou2022spikformer}, which employs a spike-based attention mechanism, achieve significantly higher accuracy but consume over ten times the energy compared to our model.

\begin{table}[h!]
\centering
\caption{The computation and estimated energy cost}
\label{tab:energy}
\resizebox{\linewidth}{!}{
\begin{tabular}{crcrrr}
\toprule Model & {\#Param.} & Accuracy[$\%$] & {\#Add.} &{\#Mult.} & {Energy}\\ 
\midrule
SparseMLP (ANN) & 24M & 81.9 & 2.50G & 2.50G & 13.75 $\mathrm{mJ}$  \\
SpikFormer & 30M & 73.38 & 11.09G & - & 11.58 $\mathrm{mJ}$  \\
Spiking ResNet-34 & 22M & 63.72 &  1.85G & 118M & 2.21 $\mathrm{mJ}$  \\
Spiking MLP-SPE-T & 25M & 66.39 &  1.18G &	12M & \textbf{1.12 $\mathrm{mJ}$}  \\
\bottomrule
\end{tabular}
}
\label{comp}
\end{table}

\subsection{Visualization}
\begin{figure*}[t!]
  \centering
  \begin{subfigure}{0.9\linewidth}
  \centering
      \includegraphics[width=1\linewidth]{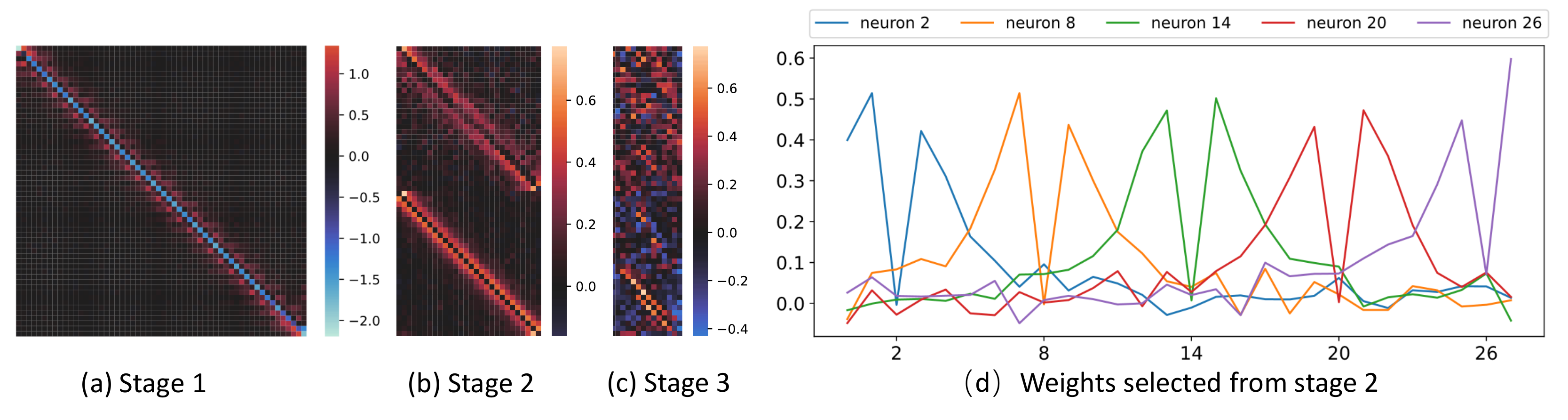}
  \end{subfigure}
  \caption{Axial sampling weights from token blocks across different stages after training on ImageNet-1K. Figures (a-c) show randomly selected token weights plotted in a $H \times H$ or $W \times W$ 2D matrix. 1, 2, and 4 weight matrices from stage 1, 2, and 3 are displayed, respectively. The diagonal distribution of values in the early stages indicates local sampling. Figure (d) represents weights of several neurons from the same token block in stage 2. For the sake of simplicity, we only plot a few neurons here, however, it should be noted that this 'M' shaped weight kernel is a universal finding in the network's early stages.}
  \label{fig:rcp}
\end{figure*}

Lastly, we provide a visualization of weights from token blocks across various stages of the spiking MLP-SPE-T after training on ImageNet-1K. These weights are depicted in Figure \ref{fig:rcp}. Interestingly, the weights in the early stages appear to concentrate on local areas, while those in the later stages progressively sample more global areas. Without an inductive bias, the SNN naturally learns to form hierarchically arranged receptive fields. Further, the receptive fields of spiking neurons in the initial stages resemble the 'off-center on-surround' receptive fields of cortex cells \cite{hubel1962receptive}, characterized by a central inhibitory area encircled by an excitatory area. This 'M' shaped weight kernel is universally observed in the network's early stages. It would be intriguing to determine whether similar phenomena occur in ANNs.

\section{Conclusion\label{sec:conclusion}}
This work presented the construction of MLP architectures utilizing spiking neurons, adhering to the MFI principle. The spiking MLP-Mixer, within our framework, employed batch normalization instead of layer normalization in both the token and channel block to ensure compatibility with MFI. Furthermore, the network's local feature learning capability was augmented with a spiking patch encoding layer, which notably enhanced the network's performance.

Leveraging these foundational building blocks, we investigated an optimal skip connection configuration and developed a proficient multi-stage spiking MLP network. This network combined global receptive field and local feature extraction. With comparable model capacity, our networks markedly surpassed state-of-the-art, mainstream deep spiking convolutional networks on the ImageNet-1K dataset. This is evident in terms of balancing accuracy, model capacity, and computation cost, thus demonstrating the efficacy of this alternative architecture for deep SNNs.

Our work underscores the importance of integrating global and local learning for optimal SNN architecture design. Interestingly, the trained receptive fields of our network bear a striking resemblance to those of cells in the cortex. Future comparative analysis with ANNs could potentially yield insightful results. In terms of accuracy, however, our SNN-based MLPs are still suboptimal compared with MLP architectures in the ANN domain. 
Considering this is an initial effort in adapting SNNs to MLP frameworks, future research into more effective structures and spike-based mechanisms, such as adaptive thresholds, is likely to enhance network performance. 
Furthermore, recent works of SNNs integrated with attention mechanisms have been applied to challenging language tasks \cite{zhu2023spikegpt, bal2023spikingbert}. 
With the ever growing demand on computational efficient architectures for these tasks, the proposed spike-based MLP structure could be integrated as a light weight module in the encoder network.

\bibliography{main} 
\bibliographystyle{IEEEtran}

\end{document}